\documentclass{article}
\usepackage{spconf,amsmath,graphicx}
\usepackage{times}
\usepackage{psfrag,pstricks,pstricks-add,pst-node}
\usepackage{graphicx}
\usepackage{papertweaker}
\usepackage{epsfig}
\usepackage{amsmath,amssymb}
\usepackage{multirow}
\usepackage{caption}
\usepackage{subcaption}
\usepackage[printwatermark]{xwatermark}

\RequirePackage{booktabs}

\newcommand{\TODO}[1]{\textcolor{red}{TODO #1}}
\renewcommand{\TODO}[1]{~}

\title{A comparative study of batch construction strategies \\
	 for recurrent neural networks in MXNet}

\makeatletter
\def\name#1{\gdef\@name{#1\\}}
\makeatother \name{{\em Patrick Doetsch, Pavel Golik, Hermann Ney}}

\address{AppTek, \\
	6867 Elm St, Set 300,
	Mclean VA 22101 United States \\
	{\small \tt \{pdoetsch,pgolik,hney\}@apptek.com}
}

\begin{document}

  \maketitle
  \begin{abstract}
  	 In this work we compare different batch construction methods for mini-batch training
  	 of recurrent neural networks. While popular implementations like TensorFlow and MXNet suggest 
  	 a bucketing approach to improve the parallelization capabilities of the recurrent training process, 
  	 we propose a simple ordering strategy that arranges the training sequences in a stochastic alternatingly sorted way. We compare our method to sequence bucketing as well as various other
  	 batch construction strategies on the CHiME-4 noisy speech recognition corpus. The experiments
  	 show that our alternated sorting approach is able to compete both in training time and recognition performance while being conceptually simpler to implement. 
  \end{abstract}

  \section{Introduction}
  	Neural network based acoustic modeling became the de-facto standard in automatic speech recognition (ASR)
  	and related tasks. Modeling contextual information over long distances in the input signal hereby showed to 
  	be of fundamental importance for optimal system performance. Modern acoustic models therefore use recurrent 
  	neural networks (RNN) to model long temporal dependencies. In particular the long short-term memory (LSTM)~\cite{hochreiter1997lstm}
  	has been shown to work very well on these tasks and most current state-of-the-art systems incorporate LSTMs
  	into their acoustic models. While it is common practice to train the
  	models on a frame-by-frame labeling obtained from a 
  	previously trained system, sequence-level criteria that optimize the acoustic model and the alignment model jointly
  	are becoming increasingly popular. As an example, the connectionist temporal classification (CTC) \cite{CTC}
  	enables fully integrated training of acoustic models without assuming a frame-level alignment to be given. Sequence-level
  	criteria however require to train on full utterances, while it is possible to train frame-wise labeled systems
  	on sub-utterances of any resolution.
  	
    Training of recurrent neural networks for large vocabulary continuous speech recognition (LVCSR)
    tasks is computationally very expensive and the sequential nature of recurrent process prohibits to
    parallelize the training over input frames. A robust optimization requires to work on large batches of utterances
    and training time as well as recognition performance can vary strongly depending on the choice of how
    batches were put together. The main reason is that combining utterances of different lengths in a mini-batch requires 
    to extend the length of each utterance to that of the longest utterance within the batch, usually by appending zeros. These zero frames 
    are ignored later on when gradients are computed but the forward-propagation of zeros through the RNN is a waste of computing power.
    
    A straight-forward strategy to minimize zero padding is to sort the utterances by length and to partition them into
    batches afterwards. However, there are significant drawbacks to this method. First, the sequence order remains constant in each epoch 
    and therefore the intra-batch variability is very low since the same sequences are usually combined into the same batch. Second, the strategy favors putting similar utterances into the same batch, since short utterances often tend to share other properties.
    One way to overcome this limitation was proposed within TensorFlow and is also used as recommended 
    strategy in MXNet. The idea is to perform a \textit{bucketing} of the training corpus, where each bucket represents
    a range of utterance lengths and each training sample is assigned to the bucket that corresponds to its length.
    Afterwards a batch is constructed by drawing sequences from a randomly chosen bucket.
    The concept somehow mitigates the issue of zero padding if suitable length ranges can be defined, while still allowing for 
    some level of randomness at least when sequences are selected within a bucket. However, buckets have to be made very large 
    in order to ensure a sufficiently large variability within batches. On the other hand, making buckets too large will 
    increase training time due to irrelevant computations on zero padded frames. Setting these hyper-parameters correctly is
    therefore of fundamental importance for fast and robust acoustic model training.
    
    In this work we propose a simple batch construction strategy that is easier to parametrize and
    implement. The method 
    produces batches with large variability of sequences while at the same time reducing irrelevant computation to a minimum.
    In the following sections we are going to give an overview over current batch construction strategies and compare them 
    w.r.t.~training time and variability. We will then derive our proposed method and discuss its properties on a theoretical
    level, followed by an empirical evaluation on the CHiME-4 noisy speech recognition task. 
   
  \section{Related Work} \label{sec:related}
  	While mini-batch training was studied extensively for feed-forward networks \cite{Li16}, authors rarely reveal the batch construction strategy they used during training when 
  	RNN experiments are reported. This is because the systems are either trained in a frame-wise
  	fashion~\cite{2016arXiv161005256X} or because the analysis uses sequences of very similar length as in \cite{PascanuMB13}. We studied in an earlier work \cite{Doetsch14} how training on sub-sequences in those cases can lead to significantly faster and often also more robust training. 
  	In \cite{Laurent16} the problem of having sequences of largely varying lengths in a batch was identified and the authors
  	suggested to adapt their proposed batch-normalization method to a frame-level normalization, 
  	although a sequence-level normalization sounds theoretically more reasonable. In \cite{Bengio15} a curriculum learning strategy is proposed where sequences follow a specific scheduling in order to reduce overfitting.
  	
  	Modern machine learning frameworks like TensorFlow \cite{tensorflow} and 
  	MXNet \cite{mxnet} implement a bucketing approach based on the lengths distribution of the sequences. In \cite{Khomenko06} the authors extend this idea by selecting optimal sequences within each bucket using a dynamic programming technique.

  \section{Bucketing in MXNet} \label{sec:bucketing}
  Borrowed from TensorFlow's sequence training example, MXNet implements bucketing by 
  clustering sequences into bins depending on their length. The size of each bin, i.e.~the span of 
  sequence lengths associated with this bin, has to be specified by the user and optimal values 
  depend on the ASR task. The sampling process can be done in logarithmic time, since for each sequence length in the training set a binary search over the bins has to be performed. 
    
  In each iteration of the mini-batch training a bucket is then selected randomly. Within the selected bucket a random span of sequences is chosen to be used as data batch. Note that this random shuffling only ensures a large inter-batch variance w.r.t.~the sequence length, while the variance 
  within each batch can be small.
  
  Bucketing is especially useful if the RNN model itself does not support dynamic unrolling and 
  is not able to handle arbitrary 
  long sequences but instead requires to store an unrolled version of the network for every possible 
  length. In those cases bucketing allows the framework to assign each batch to the shortest possible 
  unrolled network, while still optimizing the same shared weights.

  \section{Proposed Approach} \label{sec:approach}
  
  \begin{figure}
   	\resizebox{0.48\textwidth}{!}{\includegraphics{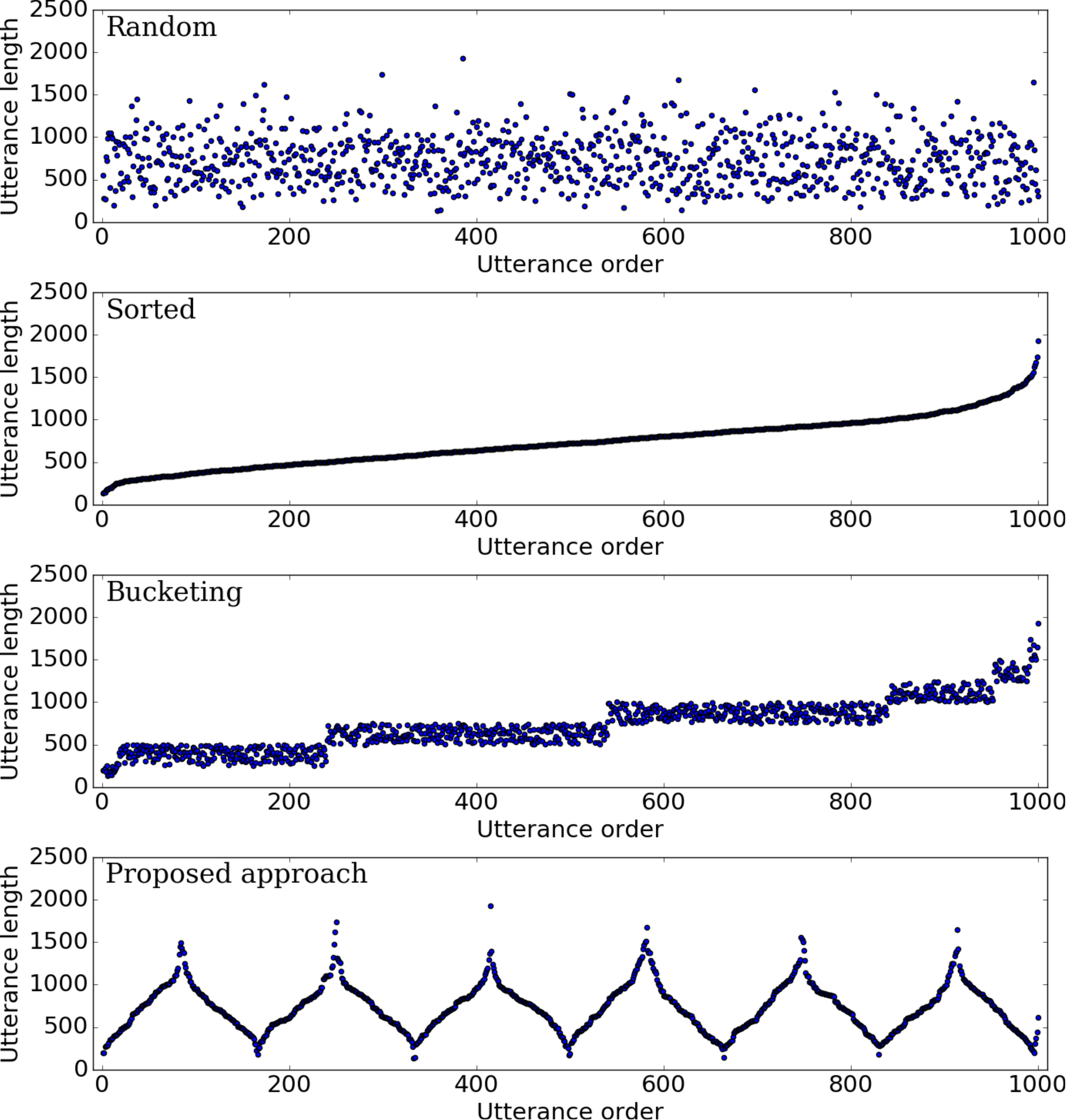}}
   	\caption{Resulting sequence ordering for different batch construction strategies. The graphs 
   	show the utternace lengths of 1000 randomly selected samples of the CHiME-4 training set. The 
   	Y-axis shows the length and the X-axis the utterance index for a given ordering. Bucketing was done with a bucket size of 250 and in the proposed approach we used 12 bins.}
   	  \label{fig:plot}
   	  \vspace{-5mm}
   \end{figure}
  
  In order to improve the intra-batch variability we propose a stochastic bucketing process. At the beginning
  of each epoch the utterances are arranged randomly and then partitioned into bins of equal size. Each bin
  is then sorted in alternating directions such that two consecutive bins are sorted in reverse order to each 
  other. Finally, the constructed ordering is partitioned into batches. The overall algorithm can be summarized as follows: \\[1ex]
 
 \begin{minipage}{\textwidth}
 	\vspace{-4mm}
 	For each epoch
 	\begin{enumerate}
 		\item shuffle training data 
 		\item partition resulting sequence into $N$ bins
 		\item sort each bin $n$ by the utterance length:
 		\begin{itemize}
 			\item in ascending order if $n$ is odd
 			\item in descending order if $n$ is even
 		\end{itemize}
 		\item draw consecutive batches of desired size \\
 		from the resulting sequence \\[1ex]
 	\end{enumerate}
 \end{minipage}

 Due to the initial shuffling and subsequent partitioning the probability for two sequences of any length being 
 put into the same bin is $\frac{1}{N\cdot(N-1)}$, so by increasing the number of bins, the variability within a 
 partition decreases quadratically while the variability among different partitions increases. The alternated
 sorting approach ensures that utterances at the boundaries of two consecutive bins are of similar length such 
 that the final partitioning into batches requires minimal zero padding.
 
 Figure \ref{fig:plot} shows the utterance lengths for random and sorted sequence ordering as well as for bucketing in MXNet and the proposed approach. Note that in the case of bucketing batches are put
 together by randomly choosing one of the buckets first, so the ordering does not directly represent 
 the final set of batches.

  \section{Experimental Setup} \label{sec:setup}
  The 4th CHiME Speech Separation and Recognition Challenge
  \cite{Vincent_CSL2016:CHiME4} consists of noisy utterances spoken by speakers in challenging 
  environments. Recording was done using a 6-channel microphone array on a tablet. The dataset revisits 
  the CHiME-3 corpus that was published one year before the challenge took place \cite{Barker2015:CHiME3}.
  
  We extracted 16-dimensional MFCC vectors as in \cite{menne16:chime4System} from the six sub-corpora
  and used them as features for the neural network models. The context-dependent HMM states were clustered into 
  1500 classes using a classification and regression tree (CART).
  We trained a GMM-based baseline model with three state HMM without skip transitions in order to obtain a frame-wise labeling of the training data. A network of three 
  bi-directional LSTM layers followed by a softmax layer was trained to minimize the frame-wise cross-entropy. Optimization was done with the Adam optimizer and a constant learning rate of 0.01. We used MXNet \cite{mxnet} for experiments using the bucketing approach and RETURNN \cite{doetsch2017:returnn} for the other methods.  
  
  After training, the state posterior estimates from the neural network are normalized by the state priors and used as likelihoods in a conventional hybrid HMM decoder using the RASR toolkit~\cite{rybach2011:rasr}.
  A 4-gram LM 
  was used during decoding with a language model scale of 12.

  \section{Experiments} \label{sec:experiments}
   In order to provide some perspective on the impact of temporal context on the CHiME-4 task, 
   we performed experiments on sub-utterance (chunk) which are presented in 
   Table~\ref{tab:chime:chunk}. For different sub-utterance lengths, we report the processing speed measured in utterances per second, the memory required and the word error rate (WER) on the evaluation set of the CHiME-4 database. Here we constrained batches to only 
   contain 5,000 frames in total, such that the overall number of updates is constant in all experiments. We can observe that 
   while large speed-ups can be obtained when training is done in this fashion, full-utterance context is
   required for optimal performance. However, it is worth noting that the memory requirement decreases 
   significantly when sub-utterance training is applied. In particular, for unusually long utterances, 
   sub-utterance training might be the only way to fit the data into GPU memory.
   
   For training on full sequences we conducted experiments with different batch construction strategies.
   The results are reported in Table \ref{tab:chime:batch}, where the first two rows show results for trivial sequence ordering 
   methods and the last rows provide a direct comparison of the bucketing approach as 
   it is implemented in MXNet and the alternated sorting approach as proposed in this paper.
    
	\begin{table}[tbp]
		\centering
		\caption{Training time and recognition performance when training is done 
		on sub-utterance level. The first column shows the maximal sub-utterance length 
		after partitioning of the original sequence. The last row shows the results 
		obtained without partitioning into sub-utterances. }
		\label{tab:chime:chunk}
		\begin{tabular}{lrrr}
			\hline
			Chunk size              & Utt./sec & Memory [GB] & WER [\%] \\
			\hline
			10     					&	36.7   & 1.6         &	21.3     \\
			50  					&	31.1   & 1.6		 &  10.1     \\
			100 					& 	29.6   & 1.6		 &	9.2      \\
			500						& 	17.3   & 1.6         &	9.0      \\
			$\max$					&	7.0	   & 5.4         &	8.9      \\
			\hline
		\end{tabular}
	\end{table}

	\begin{table}[tbp]
		\centering
		\caption{An evaluation of different sequence batch construction methods on the CHiME-4 database. Training time per epoch, memory consumption are presented in the first two columns, while the last column shows the word error rate of the corresponding acoustic model on the evaluation set.}
		\label{tab:chime:batch}
		\begin{tabular}{l@{} rrr}
			\hline
			Approach                & Utt./sec & Memory [GB] & WER [\%]          \\
			\hline
			Random 					& $\sim$ 7.0     & $\sim$ 5.4		 & $\sim$ 8.9		 \\
			Sorted					& 10.2    &	4.2 & 10.2		 \\
			\hline
			Bucketing (MXNet)		& 9.5     &	6.3		 & 9.6		 \\
			Proposed (8 bins)	    & 10.1	   & 4.8     & 9.9		 \\
			Proposed (64 bins)	    & 10.0     & 5.3     & 9.5		 \\
			Proposed (256 bins)	    & 8.8	   & 6.0     & 9.1		 \\
			\hline
		\end{tabular}
	\end{table}

  As expected, sorting the entire training set by utterance length reduces the required time 
  per epoch to a minimum, while the best overall performance is obtained when utterances are shuffled randomly. Both bucketing and the proposed 
  approach are in between. We can observe that our method is able to reach almost the same recognition performance as using a randomly shuffled sequence
  ordering, while being almost as fast as the sorted utterance scheduler. This allows for a good trade-off between runtime and system performance.

  \section{Conclusions}
    In this work we presented a novel strategy to construct sequence-level batches for recurrent 
    neural network acoustic model training. While not much attention is given to the topic 
    of batch construction, we demonstrate that different strategies can lead to large variations 
    both in training time and recognition performance. Most deep-learning frameworks rely on a bucketing approach by clustering sequences of similar length into bins and to draw batches from each bin individually. We showed that we can achieve a better runtime performance using a simpler batch design, by partitioning a shuffled sequence order and to sort the partitions in an alternating order. The method was evaluated on the ChiME-4 noisy speech recognition task and 
    compared to standard approaches like random sequence shuffling and the bucketing approach of MXNet,
    where our method was able to reach a better trade-off between training time and recognition performance while being easier to parametrize than the bucketing method.

   \section{Acknowledgements}
 
   We want to thank Jahn Heymann, Lukas
   Drude and Reinhold H\"ab-Umbach from University of Paderborn, Germany
   for their CHiME-4 front-end which we used in this work.
   
   \ninept
   \bibliographystyle{IEEEtran}
   \bibliography{strings,paper}

\end{document}